\documentclass[11pt]{article}

\usepackage[final]{acl}

\usepackage{times}
\usepackage{latexsym}
\usepackage{xspace}
\usepackage{booktabs}
\usepackage{amsmath}
\usepackage{makecell}
\usepackage{amssymb}
\usepackage{multirow}
\usepackage{enumitem}

\newcommand{\methodname}{Dude\xspace}
\usepackage[table]{xcolor}
\newcommand{\pos}[1]{\textcolor{green!50!black}{#1}}
\newcommand{\dow}[1]{\textcolor{red!70!black}{#1}}
\usepackage[most]{tcolorbox}

\usepackage[T1]{fontenc}

\usepackage[utf8]{inputenc}

\usepackage{microtype}

\usepackage{inconsolata}

\usepackage{graphicx}

\title{\methodname: A Dual-Detection Multi-Agent System for\\Paper-Code Discrepancy Detection}

\author{
 \textbf{Weijie Liu\textsuperscript{1}},
 \textbf{Running Zhao\textsuperscript{1}},
 \textbf{Wenhao Yuan\textsuperscript{1}},
 \textbf{Jinfeng Xu\textsuperscript{1}},
\\
 \textbf{Zhanfeng Xu\textsuperscript{1}},
 \textbf{Xiaoxi Zhang\textsuperscript{2}},
 \textbf{Edith Cheuk-Han Ngai\textsuperscript{1,}}\thanks{Corresponding author}
\\
 \textsuperscript{1}The University of Hong Kong,
 \textsuperscript{2}Sun Yat-sen University
\\
\small{\texttt{liuwj0817@connect.hku.hk, chngai@eee.hku.hk}}
}

\begin{document}
\maketitle
\begin{abstract}
LLM-empowered paper-code discrepancy detection has received growing concern since the scaling of research submissions exceeds the manual review capability. However, the limited context capacity and one-sided discrepancy detection of existing single-agent LLM paradigms lead to an inferior recall performance in detecting discrepancies. In this paper, we propose \textbf{\methodname}\footnote{https://github.com/VinnyLiu0817/Dude}, the first \textbf{\underline{Du}}al-\textbf{\underline{De}}tection Multi-Agent System for paper-code discrepancy detection. We discover that the granularity asymmetry of the paper-language and code-language introduces over-interpretation and over-reporting challenges in a multi-agent system design for discrepancy detection, resulting in increasing false positives. To address this,  we propose a granularity-aligned negotiation and a two-stage salience-filtering mechanism in \methodname, which effectively prevents agents from falsely reporting discrepancies. Experimental results in real-world paper-code discrepancy datasets showcase \methodname's significant recall and precision improvement by up to \textbf{22.8\%}, increasing F1 score by up to \textbf{18.7\%} compared to baseline methods.
\end{abstract}

\section{Introduction}
Paper-code discrepancy detection aims to identify inconsistencies of research claims described in the paper manuscript and code repository. This task is crucial as such inconsistencies compromise the credibility and reproducibility of research findings. Yet, the rapid growth of paper submissions has gone beyond reviewers' capacity to conduct accurate assessments of paper-code consistency within tight timelines. Therefore, leveraging the power of LLMs to conduct automated discrepancy detection has attracted increasing attention~\cite{you2026preventing, liu2026last, xu2026papers}. 

\begin{figure}
    \centering
    \includegraphics[width=1.0\linewidth]{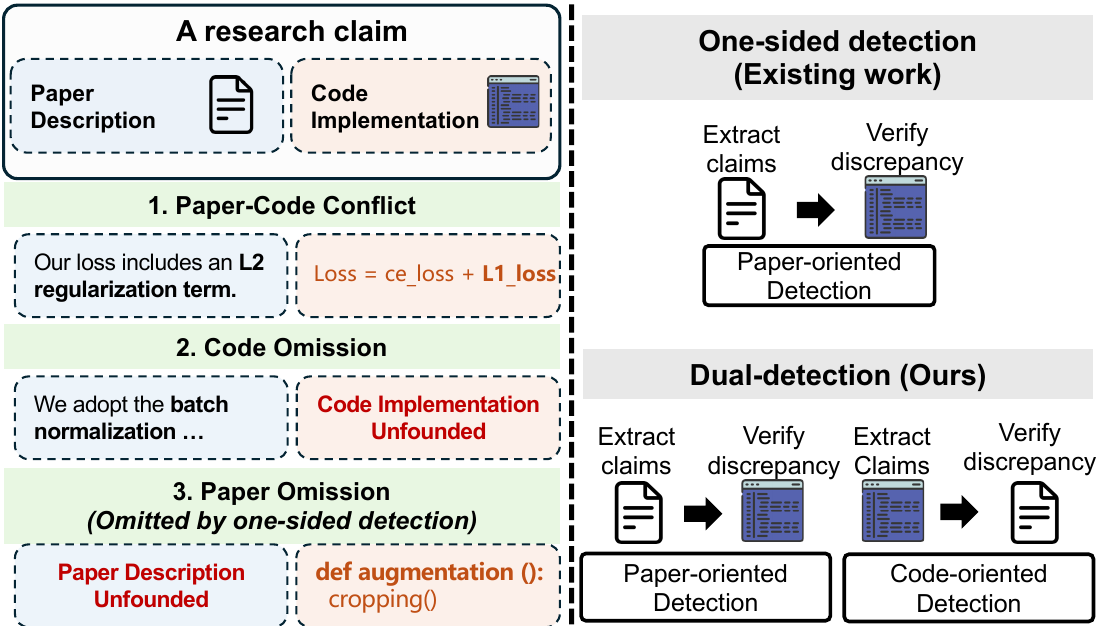}
    \caption{Three types of paper-code discrepancy of a research claim (Left) and the comparison of one-sided detection and our proposed dual-detection (Right).}
    \label{fig:pc-defintion}
    \vspace{-6mm}
\end{figure}

\noindent\textbf{Limitations.} However, a recent study ~\cite{baumgartner2026scicoqa} has reported the inferior performance of the single-agent paradigm in detecting paper-code discrepancies, with the best-performing models attaining only 46\% recall on a real-world discrepancy dataset SciCoQA. This sub-optimal performance of single-agent LLMs stem from: 1) The limited context window makes it difficult to jointly reason over lengthy papers and large codebases. 2) LLM tends to adopt paper-oriented detection by only generating a research claim from the paper, then searching the codebase for matching code. While this is effective to detect paper-code conflict and code-omission discrepancies which are explicitly stated in the paper, it is inherently limited to spot paper-omission discrepancies only observable from the code, as we illustrated in Fig.~\ref{fig:pc-defintion}.

To address these limitations, we design \textbf{\methodname}, a \textbf{\underline{Du}}al-\textbf{\underline{De}}tection Multi-Agent System for paper-code discrepancy detection. \methodname decomposes the discrepancy detection task into fine-grained subtasks and assigns specialized paper agents and code agents to conduct either paper-side understanding or code-side analysis. This alleviates the context burden of individual agents, and enables them to collectively perform both paper-oriented and code-oriented detection, i.e., dual-detection.

However, designing an effective dual-detection multi-agent system is non-trivial. The core challenge arises from \textit{the granularity asymmetry between the paper-language description and code-language implementation of a research claim}, introducing unique challenges in paper-oriented and code-oriented discrepancy detection, respectively.

\noindent\textbf{Over-interpretation in paper-oriented detection.} In paper-oriented detection, research claims extracted from papers are often condensed and high-level, leaving multiple valid code implementation choices. This ambiguity may cause the code agent to over-interpret the claim and mistakenly treat its inferred implementation as paper-grounded claims before examining the code. Consequently, this ungrounded over-interpretation leads to falsely reporting a discrepancy when the code implements one valid interpretation of a high-level research claim, but not the particular implementation that the code agent inferred without textual support from papers.

\noindent\textbf{Over-reporting in code-oriented detection.} In code-oriented detection, research claims extracted from the code are concrete and fine-grained. However, not every code implementation constitutes a research claim that requires documentation in the paper. Codebases also include environment setup and other auxiliary implementations that are necessary only for execution but peripheral to scientific contributions. A naive code-oriented detection over-reports these non-salient implementations as research claims, resulting in false-positive discrepancies.
Tab.~\ref{tab:precision_drop} shows that simply decomposing dual-detection to multiple agents increases recall, but incurs significant precision drop, showing the negative effect of over-interpretation and over-reporting.

\begin{table}[t]
\centering
\begin{tabular}{lccc}
\toprule
Method & Recall & Precision & F1 \\
\midrule
Simple-LM & 58.70\% & 91.53\% & 71.52\% \\
Vanilla-MA & 75.00\% & 74.19\% & 74.59\% \\
Improvment & \pos{+16.30 \%} & \textbf{\dow{-17.34 \%}} & \pos{+3.07 \%} \\
\bottomrule
\end{tabular}
\caption{Precision, recall, and F1 score of a single-agent method (Simple-LM) and a naive multi-agent method (Vanilla-MA), both methods adopt GPT-5.4.}
\label{tab:precision_drop}
\vspace{-6.5mm}
\end{table}

To mitigate the over-interpretation challenge, we introduce a granularity-aligned negotiation mechanism in \methodname, which enables multi-round interactions between the paper agent and code agent. During this process, the paper agent progressively refines and provides increasingly concrete and grounded paper description of the research claim. This mitigates over-interpretation by the code agent and yields more reliable discrepancy judgments. To mitigate over-reporting, \methodname introduces a two-stage filtering module that combines anchor-guided filtering with evidence-based filtering. This module suppresses research claims over-reported by code agents, thereby preventing paper agents from falsely reporting discrepancies in trivial code implementations. Our key contributions are as follows:
\begin{itemize}
    \item We reveal the limitation of single-agent LLM to detect paper-code discrepancies stems from its one-sided discrepancy detection.
    \item We design \methodname, a dual-detection multi-agent system with granularity-aligned negotiation and filtering modules, bridging the granularity asymmetry between papers and codebases.
    \item Experiments demonstrate significant recall and precision improvement of \methodname in detecting real-world paper-code discrepancies.
\end{itemize}

\section{Related Work}

\noindent\textbf{Error detection in research papers and codes.}
Leveraging the power of LLM to detect errors in research articles or open-sourced codes has attracted growing attention in current LLM era. For paper error detection, prior works include identifying logical issues~\cite{liu2023reviewergpt}, incorrect data calculations~\cite{bianchi2025err}, invalid arguments~\cite{xi2025flaws}, flawed proofs or experiment designs~\cite{zhang2025reviewing}. For code error detection, existing work covers code-comment inconsistency detection~\cite{ratol2017detecting, panthaplackel2021deep}, bug rectification~\cite{rong2025code}, reproducibility~\cite{weng2025deepscientist, bogin2024super}, and quality evaluation~\cite{tong2024codejudge}. However, these works focus either solely on paper understanding or code analysis, neglecting the potential discrepancies between a research paper and its associated codebases.

A recent study ~\cite{baumgartner2026scicoqa} presents a comprehensive analysis along with SciCoQA, a real-world paper-code discrepancy dataset, revealing the inferior performance of the latest LLMs in detecting such discrepancies. Another concurrent work BioCon~\cite{xu2026papers} also studies the paper-code inconsistencies between publications and their associated software design in the bioinformatics field. However, these two works only consider adopting single-agent paradigms to conduct discrepancy detection. The limited context capability and one-sided detection of a single LLM hinder its ability to comprehensively identify all potential inconsistencies.

\noindent\textbf{Multi-agent LLM Collaboration.} To tackle increasingly complex and long-horizon tasks, LLM-based agent systems have developed from single-agent reasoning to multi-agent collaboration~\cite{lee2025unidebugger, zhang2025orchestrating, yuan2025ma, chen2026macrollm}. In the multi-agent paradigm, a complex task is often decomposed into a set of subtasks~\cite{hong2024metagpt, liao2025agentmaster}, which are then assigned to multiple specialized agents to alleviate the context burden of individual agents. Moreover, building an effective multi-agent system typically requires a carefully designed framework to ensure efficient and desirable collaboration. Existing multi-agent collaboration frameworks are often tailored to the specific task properties like MedAgents~\cite{tang2024medagents} and ChatDev~\cite{qian2024chatdev}, or motivated by human collaboration workflows like Multi-Agent Debate~\cite{zhang2025madawsd, liang2024encouraging}. 
\section{Methodology}
\begin{figure*}
    \centering
    \includegraphics[width=1.0\linewidth]{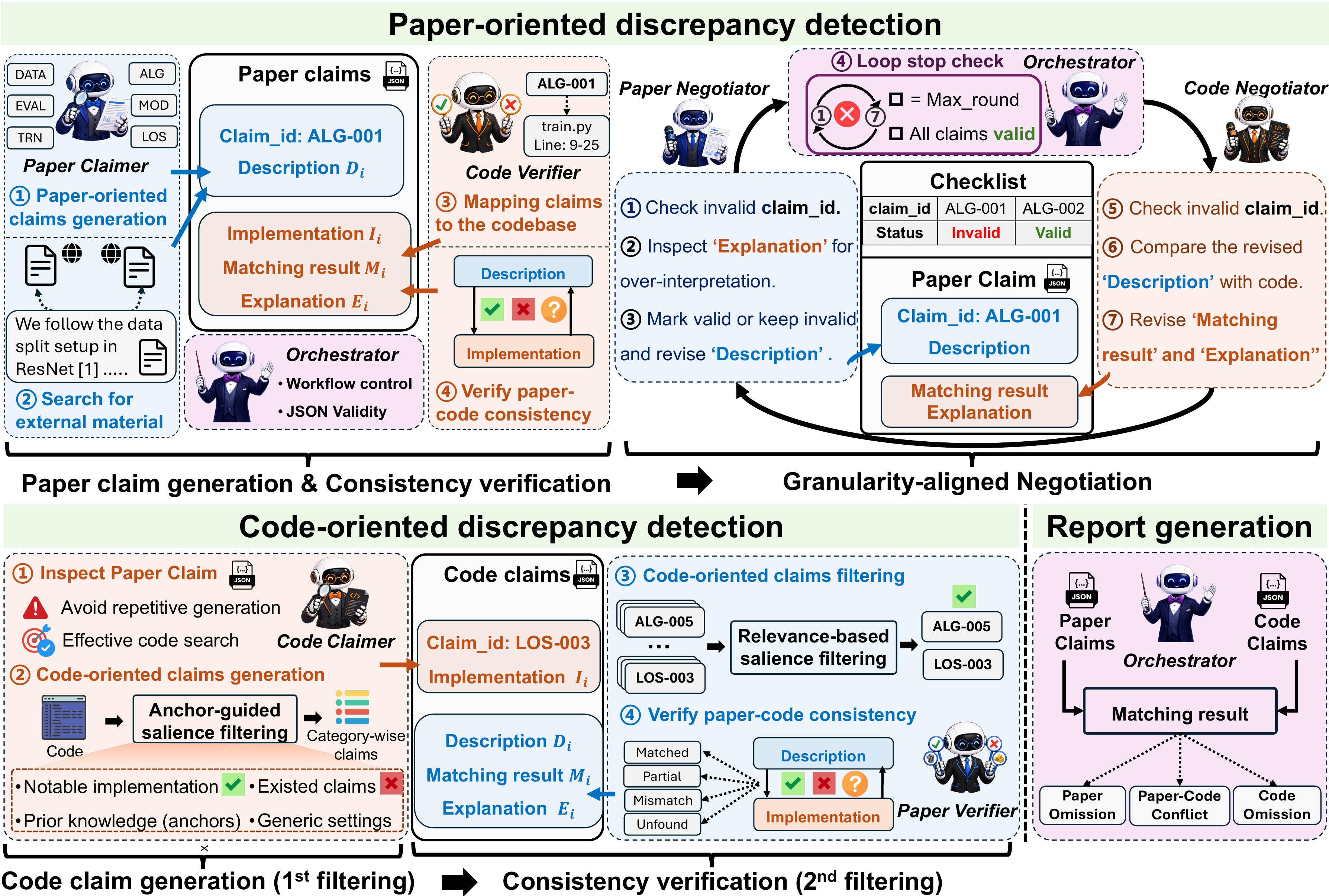}
    \caption{A complete overview of our proposed dual-detection multi-agent system \textbf{\methodname}, including three consecutive stages, paper-oriented detection (Top), code-oriented detection (Lower left), and report generation (Lower right).}
    \label{fig:full_workflow}
    \vspace{-5mm}
\end{figure*}
\subsection{Overview}
We formulate the paper-code discrepancy detection task using multi-agent LLM systems $\mathcal{M}$ as:
\begin{equation}
    \mathcal{Y} = \mathcal{M}(\mathcal{P},\mathcal{C}),
\end{equation}
where $\mathcal{P}$ and $\mathcal{C}$ denote the paper and its associated code, and $\mathcal{Y}$ represents the final discrepancy report that includes all verified research claims $R_i$ with paper-code discrepancies. For each verified research claim $R_i$, we have:
\begin{equation}
    R_i = \{D_i, I_i, M_i, E_i, O_i\},
\end{equation}
where $D_i$ denotes its paper description, $I_i$ denotes its code implementation, $M_i$ denotes the matching results between the paper and code, $E_i$ denotes the explanation of the matching result, and $O_i\in \{p,c\}$ indicates whether this research claim is initiated from the paper-oriented detection or code-oriented detection. We define $\mathcal{R}_p = \{ R_i \mid O_i = p \}$ and $\mathcal{R}_c = \{ R_i \mid O_i = c \}$ as the set of verified research claims generated in paper-oriented and code-oriented detection, referred to as paper-oriented claims and code-oriented claims. \footnote{We use `paper claim' and `paper-oriented claim' interchangeably, same for `code claim' and `code-oriented claim'.}

Our proposed \methodname includes four types of specialized agents (\textbf{claimer}, \textbf{verifier}, \textbf{negotiator}, \textbf{orchestrator}) to decompose the discrepancy detection process into three stages: paper-oriented detection (\S\ref{subsec:paper-oriented}), code-oriented detection(\S\ref{subsec:code-oriented}), and report generation(\S\ref{subsec:report-generate}), as shown in Fig.~\ref{fig:full_workflow}. In paper-oriented detection, a paper claimer initializes research claims $R_i$ by extracting paper descriptions $D_i$ from the paper, then a code verifier searches for corresponding code implementation $I_i$ to verify their consistency, presenting the matching result $M_i$ and explanation $E_i$ to form the verified paper-oriented claims $\mathcal{R}_p$. In code-oriented detection, a code claimer initializes research claims by extracting notable implementations from the codebase, then a paper verifier retrieves corresponding paper description to verify their consistency, returning the matching result $M_i$ and explanation $E_i$ to form the code-oriented claims $\mathcal{R}_c$. During this process, two negotiators refine claim description and explanation to mitigate over-interpretation, and an orchestrator coordinates workflow and consolidates $\mathcal{R}_p$ and $\mathcal{R}_c$ into a final report $\mathcal{Y}$.

\subsection{Paper-oriented Discrepancy Detection}
We formulate the paper-oriented discrepancy detection process $f_{paper}(.)$ as:
\begin{equation}
    \mathcal{R}_p = f_{paper}(\mathcal{P}, \mathcal{C}, \mathcal{A}_p),
\end{equation}
where 
$\mathcal{A}_p = \{\mathcal{A}_{pc}, \mathcal{A}_{cv}, \mathcal{A}_{pn}, \mathcal{A}_{cn}, \mathcal{A}_{O}\}$ denotes specialized agents in the paper-oriented detection.

\label{subsec:paper-oriented}
\subsubsection{Claim initialization} To begin the paper-oriented detection, a \textbf{paper claimer agent} $\mathcal{A}_{pc}$ initializes a research claim $R_i$ by extracting the paper description $D_i$ of the research claim from the paper manuscript $\mathcal{P}$:
\begin{equation}
    D_i = \mathcal{A}_{pc}(\mathcal{P}),
\end{equation}
If the research claim mentions the adoption or comparison of external research, like `Our work follows the dataset split setup in [1]...'. The paper claimer would use web search tools $\mathcal{T}_w=$\{ \texttt{WebSearch},  \texttt{WebFetch}\} to refine the description $D_i$ with fetched external knowledge $\mathcal{K}^{ext}_i = \mathcal{T}_w(\mathcal{P}, D_i)$. The paper description $D_i$ of the research claim is further refined as:
\begin{equation}
    D_i = \mathcal{A}_{pc}(\mathcal{P},\mathcal{K}^{ext}_i).
\end{equation}
\subsubsection{Paper-code consistency verification}
The \textbf{code verifier agent} $\mathcal{A}_{cv}$ then inspects the codebase $\mathcal{C}$ and maps the paper description $D_i$ to its corresponding code implementation $I_i$ using built-in code-search tools $\mathcal{T}_c=$\{\texttt{grep}, \texttt{glob}, \texttt{read}\}:

\begin{equation}
    I_i = \mathcal{T}_c(D_i, \mathcal{C}),
\end{equation}
where $I_i$ includes both the content and location of the corresponding source code. If the code implementation is not found, $I_i=\emptyset$. The code verifier $\mathcal{A}_{cv}$ then verifies the consistency between the paper description $D_i$ and code implementation $I_i$, providing the matching result $M_i$ and explanation $E_i$.
\begin{equation}
    M_i,E_i = \mathcal{A}_{cv}(D_i, I_i),
    \label{eq:verification}
\end{equation}
where $M_i\in$\{matched, partial, mismatch, unfound\}, and $E_i$ interprets the result $M_i$ based on the description $D_i$ and implementation $I_i$.

\subsubsection{Negotiation process.} Due to the granularity asymmetry between the natural language of paper description $D_i$ and code languages of implementation $I_i$, the code verifier  $\mathcal{A}_{cv}$ in Eq.~\eqref{eq:verification} may over-interpret the high-level paper description $D_i$, yielding a false matching result $M_i$ and an over-interpreted  explanation $E_i$. To this end, we design a granularity-aligned negotiation process that iteratively refines the paper description, revises explanation, and corrects matching result for all $N_p$ paper-oriented claims $\{D_i, M_i, E_i \}_{i=1}^{N_p}$. The negotiation process $f_{neg}(.)$ at the $j^{th}$ round can be formulated as:
\begin{equation}
    D_i^{j+1}, M_i^{j+1}, E_i^{j+1} = f_{neg}(D_i^{j}, M_i^{j}, E_i^{j}, \mathcal{A}), \forall i
\end{equation}
where $\mathcal{A}= \{\mathcal{A}_{pn}, \mathcal{A}_{cn}, \mathcal{A}_{O}\}$ represents the paper negotiator agent $\mathcal{A}_{pn}$, code negotiator agent $\mathcal{A}_{cn}$, and orchestrator agent $\mathcal{A}_{O}$.

As shown in the negotiation process in Fig.~\ref{fig:full_workflow}, a checklist $\mathcal{L} = \{L_{i}\}_{i=1}^{N_p}$ records the validity of each claim explanation $E_i$, where $L_{i}$ is defined as:
\[
L_{i} =
\begin{cases}
\texttt{valid},&E_i\text{ has no over-interpretation},\\
\texttt{invalid},&E_i\text{ has over-interpretation}.
\end{cases}
\]
To initiate the negotiation process, the \textbf{ paper negotiator} $\mathcal{A}_{pn}$ first inspects explanation $E_i$ of each claim to identify over-interpretation. If the explanation $E_i$ is accurate, the paper negotiator marks $L_i = \texttt{valid}$ on the checklist. For claims with over-interpreted explanation, the paper negotiator refines the description $D_i$ based on the paper $\mathcal{P}$ to resolve the extrapolation that causes over-interpretation. Formally, the refinement of paper description $D_i^j$ at the $j^{th}$ negotiation round is formulated as:
\begin{equation}
    D_i^{j+1} = \mathcal{A}_{pn}(D_i^j, E_i^j, \mathcal{P} )
\end{equation}
Next, the \textbf{code negotiator agent $\mathcal{A}_{cn}$} locates the code implementation $I_i^{j+1}$ corresponding to the refined paper description $D_i^{j+1}$, re-verifies their consistency, and updates the matching result $M_i^{j+1}$ along with a refined explanation $E_i^{j+1}$ for the examination the next negotiation round. The revision of matching result and explanation at the $j^{th}$ negotiation round is formulated as:
\begin{equation}
    M_i^{j+1}, E_i^{j+1} = \mathcal{A}_{cn}(D_i^{j+1}, I_i^{j+1})
\end{equation}
The \textbf{orchestrator agent} $\mathcal{A}_O$ coordinates the overall negotiation process, and terminates it when $L_i = \texttt{valid}, \forall i$, or when the process reaches a pre-defined maximum number of rounds $r$. At this point, our \methodname has produced a set of verified research claims $\mathcal{R}_p = \{R_i\mid O_i = p\}$ from the paper-oriented detection process.

\subsection{Code-oriented Discrepancy Detection}
\label{subsec:code-oriented}
We formulate the code-oriented discrepancy detection process $f_{code}(.)$ as:
\begin{equation}
    \mathcal{R}_c = f_{code}(\mathcal{C}, \mathcal{P}, \mathcal{A}_c),
\end{equation}
where $\mathcal{R}_c$ denotes the generated code-oriented research claims, and $\mathcal{A}_c = \{\mathcal{A}_{cc}, \mathcal{A}_{pv}\}$ denotes all specialized agents in the code-oriented detection.

\noindent\textbf{Claim generation and anchor-guided filtering.} 
In code-oriented discrepancy detection, a \textbf{code claimer agent} $\mathcal{A}_{cc}$ initializes a research claim $R_i$ by extracting the notable code implementations $I_i$ of the research claim from the codebase $\mathcal{C}$. 
 We formulate the claim generation process as:
\begin{equation}
    I_i = \mathcal{A}_{cc}(\mathcal{R}_{p}, \mathcal{C}, \mathcal{K}_a)
\end{equation}
Before examining the repository $\mathcal{C}$, the code claimer first inspects the paper-oriented claims $\mathcal{R}_{\mathcal{P}}$ to avoid extracting repetitive research claims and enable more efficient search, since $\mathcal{R}_p$ contains existing paper-oriented claims and their code implementation. To preserve only notable implementations $I_i$ and discard those trivial ones, we adopt an anchored-guided filtering module by adopting category-wise domain knowledge $\mathcal{K}_a$ as anchors to help code claimer better understand the criterion of the implementation significance from different categories. In this way, the code claimer can more accurately identify the salient code implementation $I_i$ and thus alleviate the over-reporting problem.

\noindent\textbf{Relevance-based filtering and consistency verification.}
However, the importance of code implementations is not an intrinsic property of the code alone. It is often context-dependent and paper-sensitive. The same implementation may be highly important in one type of article but peripheral in another. For instance, an INT8 quantization module is essential to report in a model compression paper, since it directly affects paper's contribution about memory usage and efficiency. However, the same quantization code is trivial in a paper about new attention architectures for classification, whose contribution has little connection with quantized models. Therefore, the \textbf{paper verifier agent} $\mathcal{A}_{pv}$ conducts a relevance-based salience filtering $f_{filter}(.)$ to remove trivial research claims and extract paper description $D_i$ of research claim whose implementation $I_i$ has high relevance to the paper $\mathcal{P}$.
\begin{equation}
    D_i = f_{filter}(\mathcal{A}_{pv}, \mathcal{P}, I_i).
\end{equation}
This prevents the paper verifier from reporting trivial inconsistencies in the verification process, where the paper verifier examines the consistency between the code implementation $I_i$ and paper description $D_i$, and returns the matching result $M_i$ and explanation $E_i$.
\begin{equation}
    M_i,E_i = \mathcal{A}_{pv}(D_i, I_i).
\end{equation}
At this point, our \methodname has produced a set of verified research claims $\mathcal{R}_c = \{R_i\mid O_i =c\}$ from the code-oriented detection process.

\subsection{Discrepancy Report Generation}
\label{subsec:report-generate}
To yield the final discrepancy report $\mathcal{Y}$, the orchestrator agent $\mathcal{A}_O$ merges the paper-oriented claims $\mathcal{R}_{p}$ and code-oriented claims $\mathcal{R}_{c}$ and determines the discrepancy type $t_i$ of research claim $R_i$ based on its matching results $M_i$ and origination $O_i$.
\[
\begin{small}
t_i =
\begin{cases}
\texttt{Conflict},&M_i \in \{\texttt{partial}, \texttt{mismatch}\},\\
\texttt{Paper-omission}, &M_i = \texttt{unfound}, O_i = c,\\
\texttt{Code-omission}, & M_i = \texttt{unfound}, O_i = p.
\end{cases}
\end{small}
\]
All \texttt{partial} and \texttt{mismatch} claims are treated as paper-code conflicts. For \texttt{unfound} claims, if the claim is paper-oriented ($O_i=p$), it indicates a code omission. Otherwise, if the claim is code-oriented ($O_i=c$), it indicates a paper omission, and all \texttt{matched} claims are excluded.

\section{Experiment}
\subsection{Experimental Setup}

\noindent\textbf{Dataset and models.} We evaluate \methodname using four closed-source and open-source LLMs (GPT-5.4, DeepSeekV4-Pro, Kimi, Claude-4.6) on the only available paper-code discrepancy dataset \textbf{SciCoQA}~\cite{baumgartner2026scicoqa}.  We also evaluate \methodname by selecting 20 research papers with public available codebases published in latest top-tier conferences (ICML 2025, ICLR 2026). 

\noindent\textbf{Baselines.} We compare \methodname against four baselines, covering both single-agent and multi-agent paradigms. For single-agent methods, we use the default method employed in SciCoQA, \textbf{Single-LM}, where a single LLM agent directly generates discrepancy report after inspecting both the paper and its corresponding code repository. We adopt \textbf{Prompt-LM} by prompting a single agent to perform paper-oriented and code-oriented detection. For multi-agent methods, we consider \textbf{Vanilla-MA} by assigning multiple agents to conduct paper-oriented and code-oriented detection. Finally, we include \textbf{Multi-Agent Debate (MAD)}~\cite{liang2024encouraging} method which aggregates multiple agents' discussion to verify paper-code discrepancies.

\noindent\textbf{Evaluation Metrics.} We follow the evaluation protocol in the SciCoQA dataset and adopt LLM-as-a-Judge~\cite{zheng2023judging} to evaluate whether the reported discrepancies match those in SciCoQA. We adopt Gemini-3.1 Pro as the judge and compute the recall, precision, and F1 scores. Notably, the reported discrepancies may include valid but unannotated discrepancies in SciCoQA. We do not count such discrepancies as true positives or false positives. We only consider discrepancies annotated in the SciCoQA dataset when computing the metrics. Nevertheless, for invalid and trivial discrepancies, we label them as false positives after verification using domain expertise and Gemini-3.1. For token consumption, we accumulate the total input and output tokens reported in the session logs or official API platforms.

\noindent\textbf{Implementation Details.} We implement \methodname on top of Codex and OpenCode~\cite{opencode2026}, enabling \methodname to be empowered by GPT, Claude, Kimi, and Deepseek. We set the default number of negotiation rounds $r=2$, and all agents employ the same LLM model unless specified. The detailed code and configurations are available at anonymous.4open.science/r/Dude

\subsection{Main Results}
\begin{table*}[t]
\centering
\setlength{\tabcolsep}{5pt}
\begin{tabular}{lcccccccc}
\hline
Method & Recall$\uparrow$ & $\Delta$R & Precision$\uparrow$ & $\Delta$P & F1$\uparrow$ & $\Delta$F1 & Token$\downarrow$ & $\Delta$Token\\
\hline

\rowcolor{gray!20}
\multicolumn{9}{c}{\textbf{GPT-5.4-xhigh}} \\
Single-LM & 58.70\% & -- & 91.53\% & -- & 71.52\% & -- & 0.71M & --\\
Prompt-LM & 64.13\% & \pos{+5.43\%} & 84.29\% & \dow{-7.24\%} & 72.84\% & \pos{+1.32\%} & 0.75M & \pos{+5.63\%}\\
Vanilla-MA & \underline{75.00\%} & \pos{\underline{+16.30\%}} & 74.19\% & \dow{-17.33\%} & \underline{74.59\%} & \pos{\underline{+3.07\%}} & \underline{1.04M} & \underline{\pos{+46.48\%}}\\
MAD & 59.78\% & \pos{+1.09\%} & \underline{93.22\%} & \pos{\underline{+1.69\%}} & 72.85\% & \pos{+1.32\%} & \textbf{1.28M} & \textbf{\pos{+80.28\%}}\\
\rowcolor{blue!10}
\methodname{} (ours) & \textbf{80.43\%} & \pos{\textbf{+21.74\%}} & \textbf{93.67\%} & \pos{\textbf{+2.15\%}} & \textbf{86.55\%} & \pos{\textbf{+15.03\%}} & 0.76M & \pos{+7.04\%}\\
\hline



\rowcolor{gray!20}
\multicolumn{9}{c}{\textbf{DeepseekV4-Pro-Max}} \\
Single-LM & 52.17\% & -- & \underline{84.21\%} & -- & 64.43\% & -- & 0.49M & --\\
Prompt-LM & 64.13\% & \pos{+11.96\%} & 77.63\% & \dow{-6.58\%} & 70.24\% & \pos{+5.81\%} & 0.59M &\pos{+20.41\%}\\
Vanilla-MA & \underline{73.91\%} & \pos{\underline{+21.74\%}} & 76.40\% & \dow{-7.81\%} & \underline{75.14\%} & \pos{\underline{+10.71\%}} & \underline{0.91M} &\underline{\pos{+85.71\%}}\\
MAD & 67.39\% & \pos{+15.22\%} & 80.52\% & \dow{-3.69\%} & 73.37\% & \pos{+8.94\%} & \textbf{0.95M} & \textbf{\pos{+93.87\%}}\\
\rowcolor{blue!10}
\methodname{} (ours) & \textbf{75.00\%} & \pos{\textbf{+22.83\%}} & \textbf{93.24\%} & \pos{\textbf{+9.03\%}} & \textbf{83.13\%} & \pos{\textbf{+18.70\%}} & 0.60M & \pos{+22.49\%} \\
\hline
\end{tabular}
\caption{The recall, precision, F1, and token usage results of \textbf{\methodname} and baseline methods on the SciCoQA dataset using GPT-5.4 and DeepseekV4 models. $\Delta$ are calculated relative to Single-LM within each model group. Bold numbers and underlined numbers denote the highest and second-highest value in each model group.}
\label{tab:main_results}
\end{table*}
\begin{figure*}
    \centering
    \includegraphics[width=1.0\linewidth]{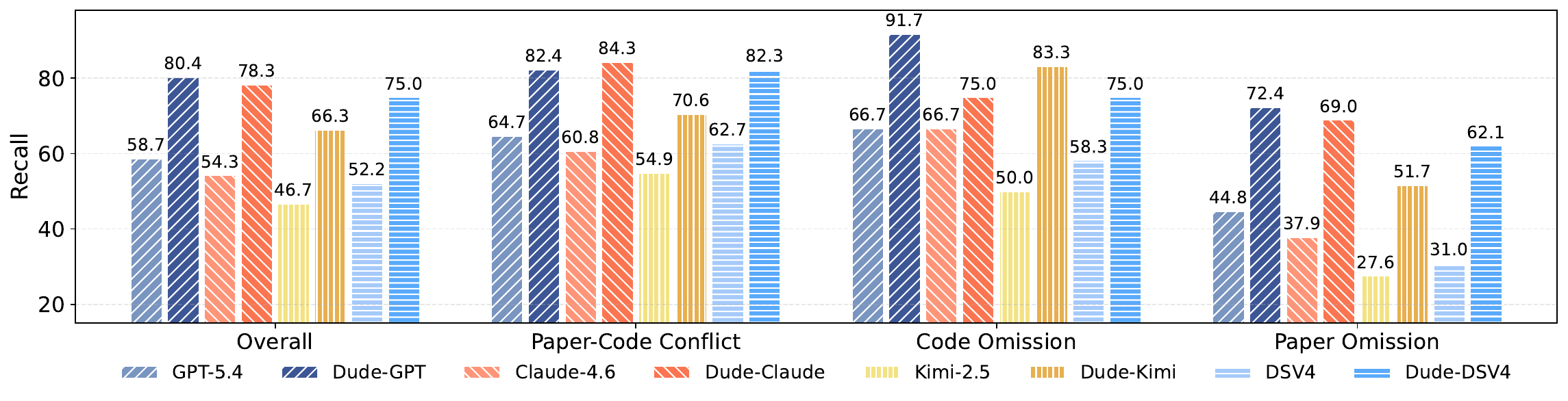}
    \caption{The recall performance of Single-LM and \textbf{\methodname} across different discrepancy types (paper-code conflict, code-omission, paper-omission) in SciCoQA dataset using GPT5.4, Claude 4.6, Kimi, and DeepseekV4 models.}
    \label{fig:recall_different_type}
\end{figure*}

\begin{table}[t]
\centering
\caption{F1 scores under different paper-side and code-side agent model combinations.}
\label{tab:mix_f1}
\begin{tabular}{lcc}
\toprule
\multirow{2}{*}{Code-side agents} & \multicolumn{2}{c}{Paper-side agents} \\
\cmidrule(lr){2-3}
 & GPT5.4 & DeepseekV4 \\
\midrule
GPT5.4      & \textbf{0.9367} & 0.9114 \\
DeepseekV4 & \underline{0.9333} & 0.9324 \\
\bottomrule
\end{tabular}
\vspace{-5mm}
\end{table}

\begin{table*}[htbp]
\centering
\begin{tabular}{lcccccc}
\hline
Method & Recall$\uparrow$ & $\Delta$R & Precision$\uparrow$ & $\Delta$P & F1$\uparrow$ & $\Delta$F1 \\
\hline

\rowcolor{gray!20}
\multicolumn{7}{c}{\textbf{GPT-5.4-xhigh}} \\
\methodname{} w/o Negotiation & 75.00\% & \dow{\textbf{-5.43\%}} & 87.34\% & \dow{-6.33\%} & 80.70\% & \dow{-5.85\%} \\
\methodname{} w/o Saliency-Filtering & 81.52\% & \pos{+1.09\%} & 74.26\% & \dow{\textbf{-19.41\%}} & 77.72\% & \dow{\textbf{-8.83\%}} \\
\rowcolor{blue!10}
\methodname{} (ours) & 80.43\% & -- & 93.67\% & -- & 86.55\% & -- \\
\hline



\rowcolor{gray!20}
\multicolumn{7}{c}{\textbf{DeepseekV4-Pro-Max}} \\
\methodname{} w/o Negotiation & 70.65\% & \dow{\textbf{-4.35\%}} & 90.28\% & \dow{-2.96\%} & 79.27\% & \dow{-3.86\%} \\
\methodname{} w/o Saliency-Filtering & 76.09\% & \pos{+1.09\%} & 81.40\% & \dow{\textbf{-11.84\%}} & 78.65\% & \dow{\textbf{-4.48\%}} \\
\rowcolor{blue!10}
\methodname{} (ours) & 75.00\% & -- & 93.24\% & -- & 83.13\% & -- \\
\hline

\end{tabular}
\caption{Ablation study of \methodname on the SciCoQA dataset using GPT5.4 and DeepseekV4. $\Delta$ values are calculated relative to \methodname{} (ours) within each model group. Bold values in the $\Delta$ column highlight the largest degradation.}
\label{tab:ablation_results}
\end{table*}
 \textbf{Overall performance analysis.} Table~\ref{tab:main_results} presents a comprehensive comparison of our proposed \methodname against baseline methods on the SciCoQA dataset, evaluated in terms of recall, precision, F1 score, and token consumption. The results demonstrate that \methodname consistently outperforms all baselines under both GPT-5.4 and DeepSeek-V4 settings, yielding substantial improvements of up to \textbf{22.8\%} in recall, \textbf{9.0\%} in precision, and \textbf{18.7\%} in F1 score.
 
 Moreover, the result also shows that Vanilla-MA, which performs paper-oriented and code-oriented detection (i.e., dual-detection) within a multi-agent framework, achieves a more substantial recall improvement than Prompt-LM, in which a single agent is prompted to perform dual-detection. This indicates that prompting a single-agent LLM is insufficient to perform effective dual-detection and thorough discovery of paper-code discrepancies. Despite its pronounced recall gains, Vanilla-MA suffers from a substantial drop in precision, whereas our proposed \methodname preserves high precision while still benefiting from the recall enhancement from the multi-agent dual-detection design. This showcases the effectiveness of the granularity-aligned negotiation process and two-stage filtering mechanism in \methodname, which jointly mitigates the over-interpretation and over-reporting challenges, thereby reducing false-positive discrepancy reports. 
 
 In terms of the token consumption, \methodname incurs the lowest additional overhead among all multi-agent methods. This efficiency stems from the design in which all paper agents and code agents in \methodname communicate exclusively through structured JSON files, substantially reducing the token cost of multi-agent interactions. In addition, \methodname maintains a record for all previously searched paper and code snippets, preventing agents from expending extra tokens on redundant searches.
 
\noindent\textbf{Recall analysis by discrepancy type.} To better understand the sources of the recall gains achieved by our multi-agent framework \methodname over the single-agent baseline Single-LM, we conduct a fine-grained analysis across discrepancy types. Fig.~\ref{fig:recall_different_type} compares the recall of \methodname and Single-LM under four LLM backbones (GPT, Claude, Kimi, and DeepSeek) across three discrepancy categories: paper-code conflict, paper omission, and code omission. The results show that \methodname consistently outperforms Single-LM across all three discrepancy types and all four LLM backbones, with average gains ranging from 19.1\% to 28.5\%. The improvement is most pronounced on paper omission (28.5\%), which is particularly notable given that this category has been identified as the most challenging for single-agent systems~\cite{baumgartner2026scicoqa}. Taken together, these results confirm that the multi-agent architecture in \methodname yields broad and robust recall improvements on the paper-code discrepancy detection task.

\noindent\textbf{Robustness under heterogeneous LLMs} We further evaluate \methodname when different LLMs are assigned to be the paper-side and code-side agents. Table~\ref{tab:mix_f1} reports F1 scores under four configurations: two hybrid settings with GPT-5.4 and DeepSeekV4 assigned to opposite sides, and two uniform settings using the same model on both sides. The results show that \methodname maintains robust discrepancy detection performance even under mixed LLM model assignments. We further observe that replacing the paper-side agents has a larger impact than replacing the code-side agents, suggesting that the capability of the paper-side agents plays a more crucial role in overall performance.

\subsection{Ablation studies} Table~\ref{tab:ablation_results} presents the ablation study of our proposed \methodname on SciCoQA dataset using GPT5.4 and DeepseekV4, evaluating the individual contributions of our designed negotiation and saliency-filtering components in \methodname. The results show that removing either one of these two components results leads to a degradation in F1 score, confirming the effectiveness of both designs in discrepancy detection. More specifically, removing the saliency-filtering module causes a significant drop in precision. Although this filtering module can occasionally filter out valid discrepancies, it is essential for suppressing the large number of trivial research claims over-reported by the multi-agent system. Removing negotiation process, on the other hand, leads to a degradation in both recall and precision, indicating the necessity of bridging the granularity gap between natural language and code language during the consistency verification process. 

\subsection{Parameter sensitivity analysis}

\begin{figure}[t]
    \centering
    \includegraphics[width=0.9\linewidth]{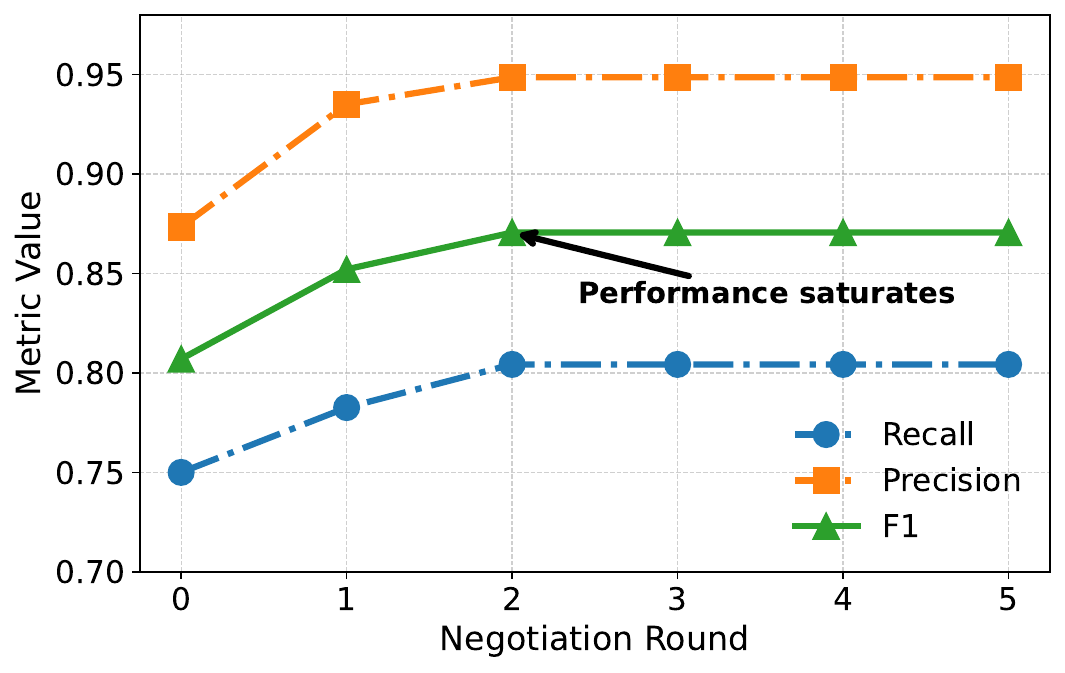}
    \caption{The effect of negotiation rounds $r$ on Recall, Precision, and F1 (\methodname-GPT-5.4)}
    \label{fig:nego_round}
\end{figure}

We investigate the impact of the number of negotiation rounds $r$ on \methodname's discrepancy detection performance. As illustrated in Fig.~\ref{fig:nego_round}, the recall, precision, and F1 score of \methodname exhibit monotonic improvement as the number of negotiation rounds $r$ increases, reaching a plateau at $r=2$, indicating that the performance saturates beyond this point. We attribute this behavior to the observation that the majority of research claims can be resolved within two rounds of discussion. As a result, even when $r$ is configured to a larger value, the negotiation process generally terminates within the first two rounds.

\subsection{Results on latest real-world research} 
\begin{table}[t]
\setlength{\tabcolsep}{2pt}
\centering
\caption{The number of valid and invalid reported discrepancies as well as the precision results of \textbf{\methodname} and baseline methods in 20 latest top-tier research works.}
\label{tab:iclr_result}

\begin{tabular}{lccc}
\toprule
Method 
& \makecell[c]{\# valid ($\uparrow$) \\ discrepancy} 
& \makecell[c]{\# invalid ($\downarrow$) \\ discrepancy} 
& Precision \\
\midrule
Single-LM   & 29 & 3  & 90.63\% \\
Prompt-LM   & 31 & 8  & 79.49\% \\
Vanilla-MA  & \textbf{46} & 21 & 68.66\% \\
MAD         & 34 & 6  & 85.00\% \\
\rowcolor{blue!10}
\textbf{\methodname (GPT)} & 45 & \textbf{3} & \textbf{93.75\%} \\
\bottomrule
\end{tabular}
\end{table}
To further assess the capability of \methodname, we employ \methodname with GPT-5.4 to detect paper-code discrepancies in 20 selected research papers and their corresponding publicly available code repositories. These papers were accepted by recent top-tier AI conferences, including ICML 2025 and ICLR 2026. As shown in Table~\ref{tab:iclr_result}, \methodname identifies the second-largest number of valid discrepancies while achieving the highest precision among all methods. Although Vanilla-MA detects the largest number of valid discrepancies, it produces the highest number of invalid ones, resulting in significantly low precision. These results demonstrate the strong potential of \methodname for comprehensive and reliable paper-code discrepancy detection in cutting-edge AI research.
\section{Conclusion}
In this paper, we present \methodname, a dual-detection multi-agent framework for paper-code discrepancy detection. Our analysis reveals that the inferior recall performance of the single-agent paradigm stems from its inherent one-sided discrepancy detection process, a limitation that cannot be overcame through simple prompting. While a naive multi-agent implementation alleviates this issue, it introduces a precision degradation arising from the granularity asymmetry between papers and code. \methodname addresses both limitations through a negotiation module that aligns cross-granularity representations via iterative agent dialogue, and a two-stage filtering module that suppresses over-reported false positives. Extensive experiments confirm that \methodname substantially improves recall while preserving precision, establishing a strong foundation for automated paper-code discrepancy detection.
\newpage

\section*{Limitations}
Despite its substantial improvements in recall and precision, \methodname has two limitations. First, its multi-agent design leads to higher token consumption than single-agent frameworks. As shown in Table~\ref{tab:main_results}, although \methodname incurs the lowest token consumption among all multi-agent methods, it still consumes more tokens than single-agent baselines (Single-LM and Prompt-LM). Second, due to the high inference cost of the latest models, we were unable to evaluate all baseline methods under the latest backbones of all LLM providers (GPT-5.5, Claude-Opus-4.7). Future work could investigate more token-efficient multi-agent frameworks and extend the evaluation to additional advanced LLM backbones to further assess the generalizability of our proposed \methodname framework.
\section*{Ethical Considerations}
Our proposed \methodname is designed to serve as a self-checking tool for researchers and as a paper-code discrepancy detection tool for reviewers to assess the reliability and reproducibility of scientific research. By providing a structured summary about potential inconsistencies between papers and their associated code, \methodname can help alleviate the reviewing burden on reviewers under growing submission volumes and compressed review cycles.

 Despite its strong empirical performance, the output of \methodname may still be incomplete or incorrect. Therefore, its outputs should be regarded as supportive evidence rather than fully automated decisions. Final judgments about the validity, quality, or reproducibility of a research work should remain under human oversight, with careful consideration of the paper, code, and experimental context.

 All experiments in this paper were conducted on publicly available datasets, and our use of models and data complies with their respective license terms.
\section*{Acknowledgments}
This work was supported by the UGC General Research Fund no. 17209822 and the Innovation and Technology Commission Fund no. ITS/383/23FP from Hong Kong.
 \section*{GenAI Usage Disclosure}
Generative AI tools were only used for typo revising to improve clarity and readability, and were not used for method design or experimental analysis in this work.

\bibliography{custom}

\appendix
\newpage

\clearpage
\section{Agent Prompts}
\label{sec:appendix}
The complete agent configuration files, along with their complete prompts, are available at anonymous.4open.science/r/Dude. We here briefly showcase the roles of different specialized agents in paper-oriented detection, code-oriented detection, and final report generation.
\subsection{Paper-oriented detection}
\begin{tcolorbox}[breakable, colback=white, colframe=blue!60!black, title=Paper Claimer Agent (Claim generation)]

\textbf{You are a paper-claim extraction agent for paper-to-code auditing.}

\textbf{Goal:}
Extract implementation-auditable research claims from the input paper file and generate the paper description of research claims in a single strict JSON object.



\textbf{Input:} <Input paper path>

\textbf{Output:} A paper-claim.json in the workspace root.


Your output must organize claims into these 6 categories:

\begin{enumerate}[leftmargin=8pt]
    \item Algorithm (Step order, Core logic)
   \item  Model (Architecture, Initialization)
   \item Loss (Loss definitions, Weighting, Coefficients)
   \item Evaluation (Evaluation logic, Metrics, Inference or validation procedure)
   \item Data (Dataset usage, Pre-processing, Augmentation, Filtering, Sampling)
   \item Training (Learning rate schedule, batch,  epochs, optimizer, mixed Precision, clipping, accumulation)

\end{enumerate}


\textbf{Extraction rules:}
\begin{itemize}[leftmargin=8pt]
    \item Extract atomic claims whenever possible. If a sentence contains multiple independently verifiable components, split them into multiple claims instead of combining them into one broad statement.
    \item Prefer implementation-relevant claims over broad summaries.
    \item Do not invent, infer, or hallucinate details that are not explicitly supported by the paper.
    \item Every claim must include at least one `paper\_description` item.
    \item If a claim mixes multiple concept, assign the claim to the category corresponding to the main implementation target.
\end{itemize}

\end{tcolorbox}

\begin{tcolorbox}[breakable, colback=white, colframe=blue!60!black, title=Paper Claimer Agent (Search external files)]

\textbf{Goal:} Your goal is to determine whether external material is needed for verification of each research claim, and if so, add concise, evidence-based external information to refine paper description.

\textbf{Inputs:}
\begin{itemize}[leftmargin=8pt]
    \item paper claims: <paper-claim.json path>
    \item Paper: <Input paper path>
\end{itemize}

\textbf{Output:} Update the paper-claim.json.





\textbf{Workflow:}

Step 1: Analyze each claim one by one. For each claim object, read its paper description carefully and decide whether the claim requires external material to verify.

A claim requires external material if:
\begin{itemize}[leftmargin=8pt]
    \item Verifying the claim requires establishing facts outside the current paper.
    \item The claim says the paper follows, adopts, uses, matches, extends, improves, or is consistent with an external source (paper, dataset, benchmark, method, implementation, setting, pre-processing rule, evaluation protocol).
    \item The claim depends on information that cannot be verified from the current paper alone.
\end{itemize}

A claim require no external material if:
\begin{itemize}[leftmargin=8pt]
    \item it merely mentions an external dataset, benchmark, metric, method adopted in this paper.
    \item it merely introduces or mentions related work without making a verifiable dependency claim.
    \item it is simply listing prior work as background.
\end{itemize}


Step 2: Locate the referenced external source.

If a claim requires external material:
\begin{itemize}[leftmargin=8pt]
\item Use the claim’s paper description to find where this claim appears in the paper.
\item Read the nearby context and corresponding reference.
\item Extract the key search clues, such as:
\begin{itemize}[leftmargin=8pt]
   \item cited paper title, citation number, and bibliography entry
   \item dataset, benchmark, website, repository name
   \item URL if present in the paper
\end{itemize}
\end{itemize}

Use these clues to search for the most appropriate external source on the web.


Step 3: Read the external source and fetch external evidence to refine paper description of the research claim.

\end{tcolorbox}

\begin{tcolorbox}[breakable, colback=white, colframe=orange!70!black, title=Code Verifier Agent]
\textbf{You are a careful code-verification agent for paper-to-code auditing.}

\textbf{Goals:} Your task is to map the paper description of each research claim to likely implementation in the code repository, and verify whether the paper description is supported by the codebase.

\textbf{Inputs:}
\begin{itemize}[leftmargin=8pt]
    \item Paper Claim: <paper-claim.json path>
    \item Code Repository: <Code Repository path>
\end{itemize}

\textbf{Output:}
- Update the paper-claim.json.



\textbf{Workflow:} For each research claim:
\begin{enumerate}[leftmargin=8pt]
    \item Search the code repository to look for likely implementation locations of the paper description of this claim.
    \item Add and write the `code\_location' field of this claim, if no candidate location is found, return an empty array.
    \item Start your code inspection from the location specified in `code\_location', this location is only the starting point, not necessarily the only relevant code. Follow related function calls, class definitions, utilities, configuration, and dependencies when necessary.
    \item Determine whether the code implementation is consistent with the paper description.
    \item For each research claim, add:
    \begin{itemize}[leftmargin=8pt]
        \item Matching result = \{matched, partial, mismatch, unfound\}
        \item Explanation
    \end{itemize}
\end{enumerate}
\textbf{Definition of matching result:}
\begin{itemize}[leftmargin=8pt]
    \item \texttt{matched}: The code implementation clearly and substantially supports the paper description.
    \item \texttt{partial}: The code implementation supports part of the claim, but not all of it; or the code implementation is weaker, narrower, conditional, or incomplete compared with the paper description.
    \item \texttt{mismatch}:
  The code implementation clearly contradicts the claim, or the code implements something materially different from what the paper description states.
    \item \texttt{unfound}:
  You cannot find relevant code implementation to verify the paper description, even after starting from code\_location and checking nearby/related code.
\end{itemize}
\textbf{Requirements for Explanation:}
\begin{itemize}[leftmargin=8pt]
    \item Explain why the code matches, partially matches, or mismatches with the paper description.
    \item Use the relevant code files and paper content to support your conclusion.
\end{itemize}

\end{tcolorbox}

\begin{tcolorbox}[breakable, colback=white, colframe=blue!60!black, title=Paper Negotiatior Agent]
\textbf{You are the paper negotiator in a two-agent negotiation workflow.}

Your job is to evaluate whether the current explanation for each invalid research claim becomes valid.

\textbf{Input:}
\begin{itemize}[leftmargin=8pt]
    \item Paper claim = <paper-claim.json path>
    \item Checklist = <checklist path>
    \item Paper = <Paper path>
\end{itemize}

\textbf{Outputs:} Update the paper-claim.json and checklist when necessary.


\textbf{Workflow:}
\begin{enumerate}[leftmargin=8pt]
    \item Inspect checklist and locate every claim whose status is `\texttt{invalid}`.
    \item For each invalid claim:
    \begin{itemize}[leftmargin=8pt]
        \item Locate the original paper context of the claim.
        \item Compare the original context against the explanation of the claim, and evaluate whether the explanation of the research claim is valid.
        \item Edit the checklist and paper description accordingly. 
    \end{itemize}
\end{enumerate}

\textbf{Criterion for explanation validity:}
\begin{itemize}[leftmargin=8pt]
    \item A `valid' explanation is when it no longer contains misunderstanding, factual mistakes, or unsupported extrapolation relative to the paper description and original paper context.
    \item An `invalid' explanation is when the explanation contains misunderstanding, factual error, or unsupported extrapolation, compared to the paper description and original paper context.
    \item When in doubt, keep the claim `invalid'.
\end{itemize}

\textbf{Editing requirement}:
\begin{itemize}[leftmargin=8pt]
    \item If you think that the explanation is `valid', update only checklist so this claim's status becomes `valid'.
    \item If you think that the explanation remains `invalid', keep its checklist status as 'invalid', and edit the paper description. The edit should consider resolving the misunderstanding, factual error, or unsupported extrapolation of existing over-interpreted explanation.
\end{itemize}
\end{tcolorbox}

\begin{tcolorbox}[breakable, colback=white, colframe=orange!70!black, title=Code Negotiatior Agent]

\textbf{You are the code negotiator in a two-agent negotiation workflow.}

Your job is to re-evaluate whether the revised paper description of the research claim is supported by the code repository.

\textbf{Inputs:}
\begin{itemize}[leftmargin=8pt]
    \item paper claims: <paper-claim.json path>
    \item Checklist = <checklist path>
    \item Codebase <Code repository path>
\end{itemize}


\textbf{Workflow:}
\begin{enumerate}[leftmargin=8pt]
    \item Read checklist and locate every invalid research claim.
    \item For each invalid claim:
    \begin{itemize}[leftmargin=8pt]
        \item Read and understand the revised papaer description of the research claim.
        \item Start your code inspection from the code\_location specified. Follow related function calls, class definitions, utilities, configuration, and dependencies when necessary.
        \item Determine whether the code implementation is consistent with the revised paper description.
        \item Revise the current matching result and explanation accordingly, so the paper negotiator can judge your updated explanation in the next negotiation round.
    \end{itemize}
\end{enumerate}
\end{tcolorbox}

\begin{tcolorbox}[breakable, colback=white, colframe=violet!50, title=Orchestrator Agent]
\textbf{You are the orchestrator agent} for a two-agent negotiation workflow involving:
\begin{enumerate}[leftmargin=8pt]
    \item  `paper negotiator` agent
    \item `code negotiator` agent
\end{enumerate}

Your job is to initialize a checklist, spawn two custom agents, coordinate the negotiation loop, and stop exactly under the defined stopping conditions. 

\textbf{Step 1:} Create checklist.json

At the beginning of the workflow, create checklist from the `paper-claim.json`:
\begin{enumerate}[leftmargin=8pt]
    \item Traverse claims in the exact same order as they appear in `claim\_verify.json`.
    \item Initialize every claim status as `invalid`.
    \item Initialize the `current\_round=0`, and read the pre-set maximum negotiation round from the configuration file.
\end{enumerate}

\textbf{Step 2:} Invoke `paper negotiator' agent

\textbf{Step 3:} Check stopping condition

After  the paper negotiator agent finishes, inspect checklist.json.

If either condition holds, stop the negotiation loop:
\begin{itemize}[leftmargin=8pt]
    \item All claims in the checklist are \texttt{valid}.
    \item `current\_round == maximum\_round`
\end{itemize}

\textbf{Step 4:} Invoke `code negotiator' agent

\textbf{Step 5:} Increment round counter

After the code negotiator agent finishes, set `current\_round = current\_round + 1`

This increment counts as one complete negotiation round.

Then return to Step 2.

\textbf{Final deliverables after negotiation:}

When the loop ends, produce:
\begin{itemize}[leftmargin=8pt]
    \item Final `checklist.json'
    \item Final `paper-claim.json' with revised paper description and explanation.
    \item A concise negotiation report including:
    \begin{itemize}[leftmargin=8pt]
        \item total completed rounds
        \item final valid claims
        \item final invalid claim ids
        \item whether the loop stopped because all claims became valid or because `maximum\_round` was reached
    \end{itemize}
\end{itemize}
\end{tcolorbox}

\subsection{Code-oriented detection}
\begin{tcolorbox}[breakable, colback=white, colframe=orange!70!black, title=Code Claimer Agent]
\textbf{You are a code-claim extraction agent for code-to-paper auditing.}

\textbf{Goal:} Your job is to extract auditable research claims from  the target code repository and summarize notable code implementations of the research claim in a strict JSON object.

\textbf{Input:}
\begin{itemize}[leftmargin=8pt]
    \item paper claim = <paper-claim.json path>
    \item Codebase = <Code Repository Path>
\end{itemize}

\textbf{Output:} A code-claim.json in the workspace root.

\textbf{Workflow:} 

\textbf{Step 1:} Inspect `paper-claim.json` as prior knowledge so that you can:
\begin{itemize}[leftmargin=8pt]
    \item understand the paper, repository, discrepancy categories, and already-covered paper claims.
    \item build a roadmap of likely code locations.
    \item avoid repeating claims that are already present in `paper-claim.json`;
\end{itemize}

\textbf{Step 2:} Inspect the code repository strictly one category at a time, in the following order (Algorithm, Model, Loss, Evaluation, Data, Training)

\textbf{Step 3: }For each category:
\begin{enumerate}[leftmargin=8pt]
\item Inspect the code repository and identify notable and  implementation.
\item Exclude any finding that is already covered in `paper-claim.json`, or repetitive in current 'code-claim.json'.
\item Write the code implementation as a claim for this category into `code-claim.json`.
\end{enumerate}

\textbf{Anchor-based salience filtering.}
The representative examples below are high-priority retrieval anchors. Use them to guide where you look first and what kinds of repository details deserve extra scrutiny. However, they are not exhaustive and must not limit your search space. You must still record any other code-grounded, non-redundant, impactful implementation detail even if it does not match these examples.

For the Algorithm category:
\begin{itemize}[leftmargin=8pt]
\item Hyper-parameters, thresholds, script-level settings, dynamic coefficients, or weighting decay, caching strategies that affects algorithm's step order, control flow, and core logic.
\item Code patches or fallback branches for corner cases. E.g., exception handling, fallback behavior, tie-breaking, or branch-specific shortcuts.
\end{itemize}

For the Model category:
\begin{itemize}[leftmargin=8pt]
\item Architectural model implementation details like nonlinearities, normalization, pooling, concatenation, masking, routing, gating, reshaping, residual mixing, or feature combination choices.
\item Activation functions, normalization procedures, masking policies, feature fusion details, or tensor manipulation choices that alter how representations are formed or propagated;
\end{itemize}

For the Loss or Training category:
\begin{itemize}[leftmargin=8pt]
\item Additional objective terms and regularizers to stabilize model training. E.g., code includes extra loss terms, penalty terms, priors, or fallback objectives.
\item Hyper-parameters, thresholds, script-level settings, dynamic coefficients, weighting decay, curriculum schedules, caching strategies, optimizer / scheduler details, or numerical stabilizers that affect training dynamics.
\end{itemize}

For the Evaluation category:
\begin{itemize}[leftmargin=8pt]
\item Evaluation protocol, benchmark-specific handling, scoring rules, thresholds, sample filters, or subset selection that affects the evaluation process and success criterion.
\item Exclusions, special-case metric handling, dataset-dependent or benchmark-dependent evaluation branches, or evaluation protocol shortcuts that change reported outcomes;
\end{itemize}

For the Data category:
\begin{itemize}[leftmargin=8pt]
\item Data preprocessing strategy like augmentations, truncation, clipping, subsampling feature smoothing, transformation, and prompt engineering that changes the data input distribution
\end{itemize}

\end{tcolorbox}

\begin{tcolorbox}[breakable, colback=white, colframe=blue!60!black, title=Paper Verifier Agent]
\textbf{You are a careful paper-verification agent for paper-to-code auditing.}

\textbf{Goal:} Your task is to remove trivial claims in `code-claim.json' based on their relevance to the paper, and determines whether the code implementation is supported by the paper.

\textbf{Input:}
\begin{itemize}
    \item code claims = <code-claim.json path>
    \item paper = <paper path>
\end{itemize}

\textbf{Output:} Update the code-claim.json.

\textbf{Workflow:}

For each research claim in the code-claim.json:

1. Inspect the paper carefully to look for corresponding paper description for the code implementation of the research claim.

2. Remove the trivial research claims by assessing the relevance between the code implementation and the paper contributions.

3. Determine whether the paper description matches, partially matches, mismatches, or fails to mention the code implementation described in the research claim.

4. For each research claim, add:
    \begin{itemize}[leftmargin=8pt]
        \item Matching result = \{matched, partial, mismatch, unfound\}
        \item Explanation
    \end{itemize}

\textbf{Definition of `matching result'}
\begin{itemize}[leftmargin=8pt]
    \item \texttt{matched}: The paper description clearly and substantially supports the code implementation.
    \item \texttt{partial}: The paper description supports part of the code implementation, but not all of it; or the code implementation is weaker, narrower, conditional, or incomplete compared with the paper description.
    \item \texttt{mismatch}:
  The paper description clearly contradicts the code implementation, or the code implements something materially different from what the paper description states.
    \item \texttt{unfound}:
  You cannot find relevant paper description to verify the code implementation.
\end{itemize}

\textbf{Requirements for Explanation:}
\begin{itemize}[leftmargin=8pt]
    \item Explain why the paper description matches, partially matches, or mismatches with the code implementation.
    \item Use the original paper text to support your conclusion.
\end{itemize}
\end{tcolorbox}

\subsection{Final report generation}
\begin{tcolorbox}[breakable, colback=white, colframe=violet!50,, title=Orchestrator Agent]
Your task is to read paper-claim.json and code-claim.json, and then conduct the paper-code discrepancy classification based on the matching result of research claims.

\begin{itemize}[leftmargin=8pt]
  \item For each research claim in the paper-claim.json,
  \begin{itemize}[leftmargin=8pt]
    \item If the \texttt{'matching result'} is \texttt{'matched'}, do not include.
    \item If the \texttt{'matching result'} is \texttt{'partial'} or \texttt{'mismatch'}, append the claim to \texttt{'paper-code conflict'}
    \item If the \texttt{'matching result'} is \texttt{'unfound'}, append the claim to \texttt{'code omission'}
  \end{itemize}
  \item For each research claim in the code-claim.json,
  \begin{itemize}[leftmargin=8pt]
    \item If the \texttt{'matching\_status'} is \texttt{'matched'}, do not include it.
    \item If the \texttt{'matching result'} is \texttt{'partial'} or \texttt{'mismatch'}, append the claim to \texttt{'paper-code conflict'}
    \item If the \texttt{'matching result'} is \texttt{'unfound'}, append the claim to \texttt{'paper omission'}
  \end{itemize}
\end{itemize}
\end{tcolorbox}
\section{Additional implementation details}
\subsection{Baseline method configurations}
\noindent\textbf{Single-LM:} For the Single-LM baseline method, we adopt the same prompt and configurations used in SciCoQA~\cite{baumgartner2026scicoqa}, thereby maintaining consistency with prior work and ensuring a fair comparison.

\noindent\textbf{Prompt-LM:} For the Prompt-LM baseline method, we enable a single agent to perform both paper-oriented discrepancy detection and code-oriented discrepancy detection. The prompt is revised based on the one used in Single-LM~\cite{baumgartner2026scicoqa}.
\begin{tcolorbox}[breakable, colback=white, colframe=gray, title=Prompt-LM]
You are an expert in analyzing research papers and their corresponding code implementations.
Your task is to carefully identify concrete discrepancies between what is described in a paper and what is actually implemented in the code.

\#\# What counts as a discrepancy

- A concrete paper–code discrepancy means a mismatch between what is stated in the original paper (e.g., formulas, algorithms, logic, methods, processes, or other settings) and what is implemented in the original code repository.

- Each distinct mismatch should be reported as a separate item.

\#\# Important

You should conduct both paper-oriented detection and code-oriented detection to discover the paper-code discrepancies. More specifically, you should first extract the research claim from the paper and inspect its corresponding code implementations to verify the consistency. Then you should extract the research claim from the code repository and inspect the corresponding paper content to verify the consistency.

\end{tcolorbox}

\noindent\textbf{Vanilla-MA:} We employ Vanilla-MA by removing the negotiation process and two-stage salience filtering modules in our propose \methodname. 

\noindent\textbf{MAD:} We adopt the default prompts and configurations in ~\cite{liang2024encouraging} to perform the discrepancy detection. We set the number of debaters $n_{debate}=2$ in our experiments. 

Notably, for fair comparison, all baseline methods and \methodname have the same access to the web-search tools (\texttt{WebSearch}, \texttt{WebFetch}), code-search tools (\texttt{Grep}, \texttt{Glob}, \texttt{Read}), and pdf-extraction tools (MinerU, PyMuPDF) in all experimental settings.

\subsection{Evaluation Setup}
To evaluate whether the reported paper-code discrepancies matches the discovered discrepancies in SciCoQA, we employ the same LLM-as-a-Judge protocol and prompt proposed in the original SciCoQA paper~\cite{baumgartner2026scicoqa}.
\begin{tcolorbox}[breakable, colback=white, colframe=gray, title=Prompt for LLM-as-a-Judge]
Your task is to evaluate whether a reference paper-code discrepancies matches a predicted paper-code discrepancy. Follow these steps:
    
1. Analyze which part of the paper or code each discrepancy is describing. Extract the core claims and issues from the reference and predicted discrepancies.

2. Analyze whether the core claims are about the same issue, i.e. if they describe the same or different paper-code discrepancies. The two discrepancies might use different wording or one might be more detailed than the other. Focus on whether the issue is the same, even if minor
details are different. However, if they describe different issues (even about the same topic or part of the paper or code) they do not match.

3. Provide a brief explanation of your reasoning.

\# Answer Format
Provide your answer in the following format:
<yes | no >
<Brief explanation>
\end{tcolorbox}
\subsection{Annotation and Validation Protocol}
\label{subsec:annotate}
We construct the annotations by adopting the annotation process, validation protocol, and prompt in SciCoQA. This keeps our annotation and validation consistent with this prior work while enabling a fair comparison across compared methods. The annotation and validation processes are as follows.

All compared methods, including Single-LM, Prompt-LM, Vanilla-MA, MAD, and Dude, are first applied to the 20 paper-code pairs to detect candidate discrepancies. Then we aggregate all reported discrepancies and validate them using Gemini-3.1-Pro as an LLM-as-a-Judge. When the judge's decision is ambiguous or when different LLM outputs yield conflicting evidence, authors with relevant domain expertise manually verify the discrepancy by inspecting both the paper claim and the corresponding code. 

After validation, for each valid discrepancy, Gemini-3.1-Pro is prompted to generate a standardized description of 3–5 sentences covering the paper statement, the code implementation, and where the discrepancy lies. We use these standardized descriptions as the reference annotations for all validated paper-code discrepancies identified in the 20 real-world papers.

\section{ Token Consumption Analysis}
\begin{figure}[t]
    \centering
    \includegraphics[width=1.0\linewidth]{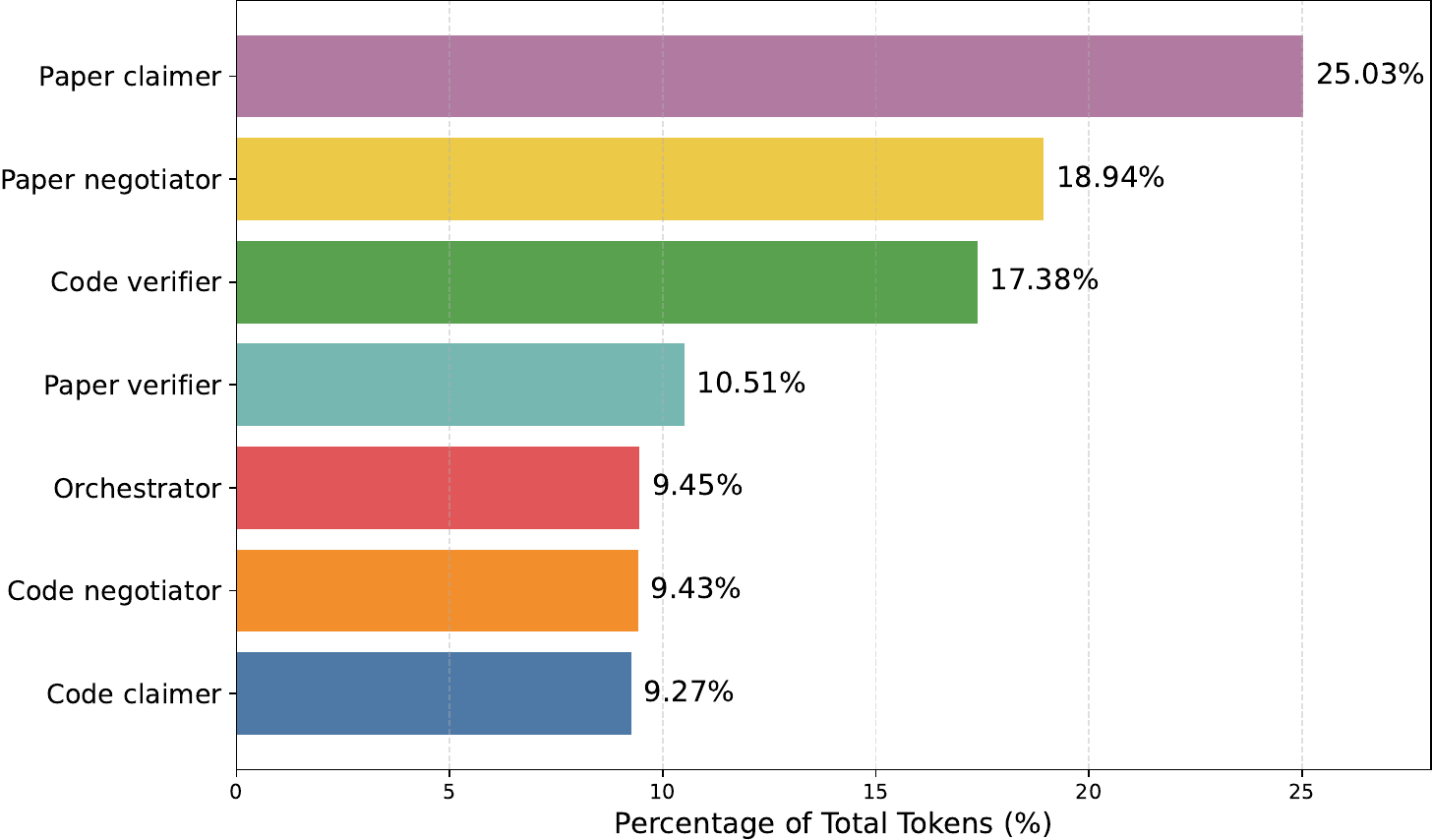}
    \caption{Token consumption percentage by agents}
    \label{fig:token-agent}
\end{figure}

\begin{figure}[t]
    \centering
    \includegraphics[width=1.0\linewidth]{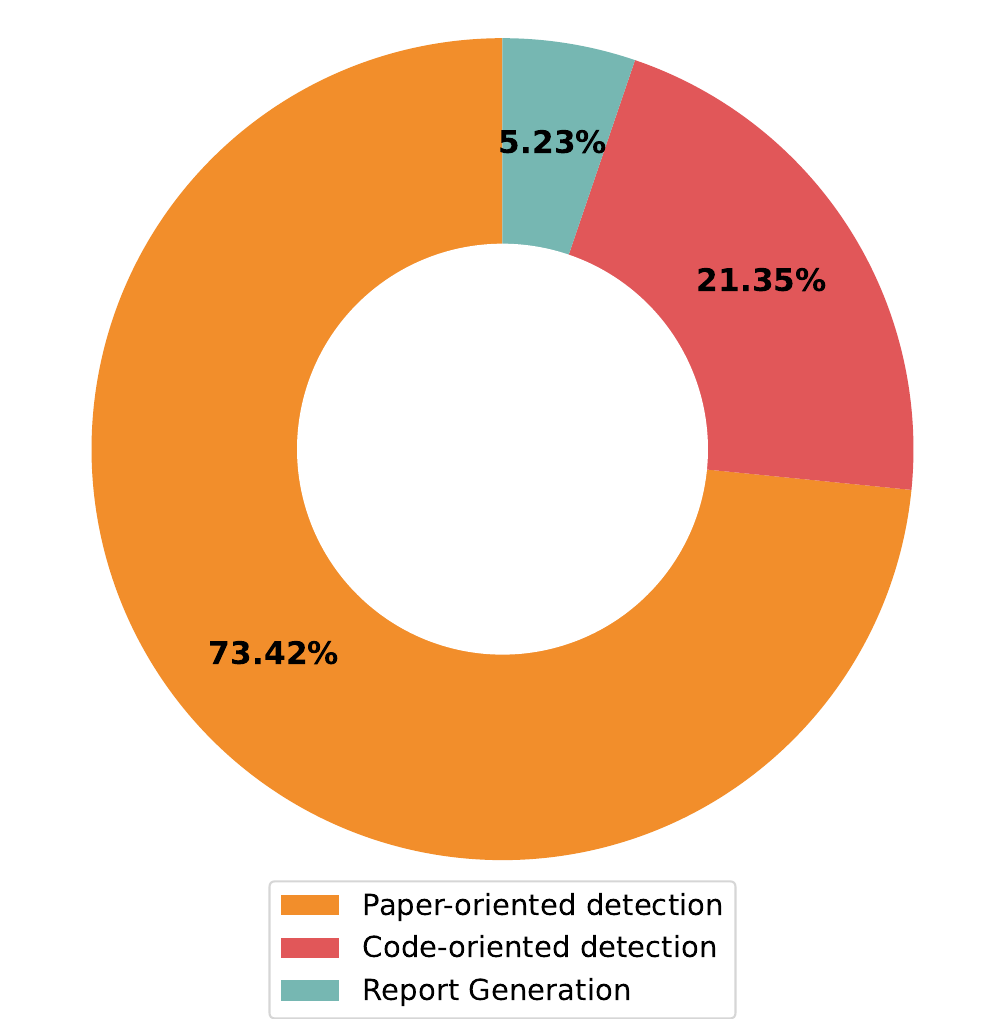}
    \caption{Token consumption percentage by discrepancy detection process}
    \label{fig:token-process}
\end{figure}
We provide a fine-grained token consumption analysis of our proposed \textbf{\methodname} by agents (Fig.~\ref{fig:token-agent}) and by the discrepancy detection process (Fig.~\ref{fig:token-process}). Fig.~\ref{fig:token-agent} presents the breakdown of token consumption across all agents, measured as the percentage of the total token usage. The results show that the paper claimer agent, paper negotiator agent, and code verifier agent are the three most token-intensive agents. Notably, the paper claimer agent accounts for nearly 25\% of the overall token consumption. This huge token consumption of paper claimer agent primarily arises from its responsibility to search for and retrieve external materials during paper-oriented generation, which substantially increases its context length and leads to markedly higher token usage compared with the other agents. The higher token consumption of paper-side agents (paper claimer, paper negotiator) also supports our empirical results in Table~\ref{tab:mix_f1}, which suggests that the capability of the paper-side agents plays a more crucial role than code-side agents in overall performance.

We also examine the token consumption of different stages in our proposed \methodname framework, including paper-oriented detection, code-oriented detection, and final report generation. Fig.~\ref{fig:token-process} shows that paper-oriented detection is the primary source of token consumption, accounting for over 73\% of the total usage and substantially exceeding the other two stages. This high cost can be attributed to two main factors. First, as we discussed above, the paper claimer agent in paper-oriented detection incurs considerable token overhead due to its need to search for and retrieve external materials for claim analysis. Second, the granularity-aligned negotiation process in paper-oriented detection requires multi-round interactions among agents, which also further increases the token budget required by the paper-oriented detection.

\section{Additional Experimental Results}
\subsection{Real-world Research Results}
Table~\ref{tab:iclr_result} reports precision on the 20 recent papers because, unlike SciCoQA, these 20 newly published works have no pre-existing discrepancy annotations to compute the recall metric. To address this, we compute recall and F1 against the reference set of 55 verified discrepancies whose annotation and validation process is detailed in Section~\ref{subsec:annotate}.

We provide the complete real-world evaluation results below, including recall, precision, F1, and average token consumption.

\begin{table}[htbp]
\begin{small}
    \centering
    \begin{tabular}{lcccc}
        \toprule
        \textbf{Method} & \textbf{Recall} & \textbf{Precision} & \textbf{F1} & \textbf{Token} \\
        \midrule
        Single-LM   & 52.73\% & 90.63\% & 66.67\% & 0.84M \\
        Prompt-LM   & 56.36\% & 79.49\% & 65.96\% & 0.91M \\
        Vanilla-MA  & 83.64\% & 68.66\% & 75.41\% & 1.33M \\
        MAD         & 61.82\% & 85.00\% & 71.58\% & 1.58M \\
        \textbf{Dude} & 81.82\% & 93.75\% & 87.38\% & 0.98M \\
        \bottomrule
    \end{tabular}
\end{small}
\caption{The recall, precision, F1, and token usage results of \textbf{\methodname} and baseline methods on the 20 real-world research using GPT-5.4.}
\label{tab:full-real}
\end{table}

Table~\ref{tab:full-real} shows that \methodname improves real-world paper-code discrepancy detection not merely by increasing the number of reported discrepancies, but by substantially improving recall while maintaining high precision. 
Notably, since the reference set is built from discrepancies reported by these methods and then validated, it may not include discrepancies that all methods missed. The recall values in Table~\ref{tab:full-real} are therefore best read as a fair comparison across methods rather than an exact measure of how many true discrepancies exist.

\subsection{Evaluation Results on GPT-5.5}
We conduct an additional evaluation on SciCoQA dataset using the latest GPT-5.5 backbone, comparing Dude against all baseline methods.
\begin{table}[htbp]
    \centering
    \begin{small}
    \begin{tabular}{lcccc}
        \toprule
        \textbf{Method} & \textbf{Recall} & \textbf{Precision} & \textbf{F1} & \textbf{Token} \\
        \midrule
        Single-LM   & 64.13\% & 93.65\% & 76.13\% & 0.79M \\
        Prompt-LM   & 64.13\% & 86.76\% & 73.75\% & 0.85M \\
        Vanilla-MA  & 73.91\% & 79.07\% & 76.40\% & 1.39M \\
        MAD         & 65.22\% & 95.24\% & 77.42\% & 1.41M \\
        \textbf{Dude} & 82.61\% & 96.20\% & 88.89\% & 0.96M \\
        \bottomrule
    \end{tabular}
    \end{small}
    \caption{The recall, precision, F1, and token usage results of \textbf{\methodname} and baseline methods on the SciCoQA dataset using GPT-5.5.}
    \label{tab:gpt5.5}
\end{table}

Table~\ref{tab:gpt5.5} shows that most methods benefit from the stronger GPT-5.5 backbone compared with GPT-5.4, indicating that the newer LLMs improve paper-code discrepancy detection. However, simply replacing the backbone with GPT-5.5 does not eliminate the limitations of the single-agent paradigm in discrepancy detection. The Single-LM baseline with GPT-5.5 still achieves only 64.13\% overall recall, while our proposed Dude continues to achieve strong performance under GPT-5.5, improving recall by 18.48\% and F1 by 12.76\% compared with Single-LM. This demonstrates that the dual-detection design and negotiation/filtering mechanisms in Dude remain effective on more recent LLM backbones.

\subsection{Human and cross-LLM Agreement}
To assess the reliability of our adopted Gemini-3.1-Pro as an LLM judge, we sample 50 paper-code discrepancy cases spanning both SciCoQA and recent real-world papers. These cases were manually reviewed by the authors with relevant domain expertise, since validating paper-code discrepancies requires understanding both paper claims and code implementation. We compare the human judgments with those produced by Gemini-3.1-Pro. Among the 50 cases, 47 received the same judgment from both, yielding a 94\% agreement rate. We also compare Gemini-3.1-Pro against a second LLM judge, Qwen-3.6. On the same 50 cases, 44 received the same judgment, yielding an 88\% agreement rate. This indicates that our LLM-as-a-Judge protocol is reasonably reliable for our evaluation, while we acknowledge that subtle cases still benefit from human evaluation.

\subsection{Representative Failure Case Study}

Although \methodname substantially improves recall and precision over existing paper-code discrepancy detection methods, it is not perfect. Our inspection shows that \methodname’s remaining false negatives mainly arise from implicit discrepancies, where the inconsistency is not stated through explicit keywords but is induced by code behavior and requires deeper reasoning to detect. On the other hand, the remaining false positives arise from ambiguous engineering code. We provide two representative failure cases of \methodname below, one false positive and one false negative, which give a clearer understanding of Dude's remaining limitations.

\noindent\textbf{Case 1 (False Positive / Incorrect Report):} Some repositories contain demonstration code in .ipynb notebooks. These notebooks may use settings that differ from the official implementation and the paper, such as running only a single trial instead of repeating experiments and reporting averaged results. These notebooks sometimes are not clearly named or commented as 'example' or 'demonstration' code. In such cases, Dude treats them as authoritative implementations and reports the difference as a discrepancy.

\noindent\textbf{Case 2 (False Negative / Missed Discrepancy):} In one case, the code restricts the outputs of the proposed module to the range (0,1) using a sigmoid activation, while the paper formulation describes these parameters as outputs of standard unbounded MLPs. Dude missed this discrepancy because the inconsistency is implicit in the activation function and requires reasoning about the numerical range induced by the code, rather than simply matching a stated paper claim to a code fragment.

\end{document}